# Pattern Recognition Theory of Mind

Gilberto de Paiva
gilbertodpaiva@gmail.com
http:www.gpaiva.com

ABSTRACT
I propose that pattern recognition, memorization and processing are key concepts that can be a principle set for the theoretical modeling of the mind function. Most of the questions about the mind functioning can be answered by a descriptive modeling and definitions from these principles. An understandable consciousness definition can be drawn based on the assumption that a pattern recognition system can recognize its own patterns of activity. The principles, descriptive modeling and definitions can be a basis for theoretical and applied research on cognitive sciences, particularly at artificial intelligence studies.

**Introduction**
The study of the mind needs overall accepted scientific basis from natural physical basis. This article proposes the fundamental principles to build a scientifically robust theory of mind based on the widely known mechanisms of pattern recognition. **I propose that the concepts of pattern recognition, pattern memorization, instinct patterns and pattern processing is a set of key scientific principles to understand the mind functioning**.

To demonstrate the scientific value of the proposed principles, I also propose a descriptive picture for the functioning of the human, animal, and artificial mind, developed from these principles. The theory includes a **comprehensive consciousness definition from physical basis**, one of most controversy scientific questions ever. The consciousness definition proposed in this theory is based on the assumption I developed that, **a mind modeled as pattern recognition system can recognize its own pattern of mental activity**. Other definitions and descriptive solutions to other fundamental questions on cognitive sciences known as easy and hard problems are also discussed.

The mind functioning principles proposed here brings is a serious attempt to answer the fundamental scientific questions on the field of cognitive sciences, philosophy of mind, artificial intelligence, biology of mind, and other related areas of mind and brain research. This proposal is scientifically important in the effort to set a scientific foundation on the mind research from the theoretical point of view, since actually no physical concept is predominantly accepted as the basis for the cognitive science. The concepts and definitions proposed here can be useful as an anchor model to others theoretical and applied studies which actually have little theoretical support to base their work.

**Fundamental Principles** and mechanisms of the functioning of the mind:

**1 – Mental Patterns**
The functionality of pattern recognition is actually a scientifically accepted mechanism that performs important role in biological neural systems, and in parallel is actually a very established computer application been a top area at artificial intelligence research. So, the manifestation, existence and functionality of the pattern recognition mechanism is a very established scientific fact.

There is no serious doubt that the brain has the ability to recognize patterns, react to patterns, remember patterns, repeat patterns, associate patterns, etc. Conceptually, experiments and scientific observations with human and animals since Pavlov and Hebb until today, and several models from artificial neural mechanism to neural networks, and even traditional computer artificial intelligence also show, incorporate, work or implement the concept of pattern recognition and processing. So, the aspect and property of pattern recognition is a very established concept about the mind, generally related to the concept of perception.

Specialized publications on neurobiology and psychology[1] and theoretical and applied recognition systems[2] shows how active is the research on the field of pattern recognition. Various scientific studies on mind research works with the concept of pattern recognition or deals with it through their argumentation in a very broad sense. But none yet took the mechanism or the concept as a fundamental explanatory basis for the functioning of the mind.

I therefore go further, and propose that **pattern recognition is a fundamental scientific principle to build a theoretical model of** mind based on natural physical grounds. So that we can explain the mind main function as to process the recognized patterns, also with memorized patterns and instinctive patterns. I will try to convince you here that this is a theoretical modeling principle valuable enough to building a theory of human, animal or artificial mind.

As patterns are recognized, memorized and processed by the mind, we can propose the concept of MENTAL PATTERN as the functional unity concept of the pattern recognition principle. The term MENTAL PATTERN encloses any pattern identifiable by any language, theory, process, mechanism, body, model, description, activity, etc. It may be a standard visual image, tactile sensation, chemical taste perception, perceived action, thought, concept, etc. I am proposing that all the humans and other animals mentally do is to process patterns.

The biological basis of the concept of mental pattern used here is in agreement to the standard bio-physical state view of the nervous system. It is well known that the nervous system of animals has activity when stimulated externally, or by physiological process even on thought and introspection (if human). At a bio-physical perspective I propose that there is a correlation between the endless possible bio-physical states of the nervous system with the endless real patterns that the mind can interact and recognize from the world included the patterns of its self functioning. This correlation process between the mental patterns and the world patterns is what could be called the mind representational framework. Thus a mental pattern (neural-physiological state) can be activated by a detected sensorial stimulus from an external (or self) physical pattern. A mental pattern may activate others mental patterns (or combination of patterns), and mental patterns may cause physiological reactions such as muscle movements, and even thinking and consciousness behavior.

**2 – Instinctive Mental Patterns**
Some patterns such as hungry, pain, sleepiness, etc, are the basic mental patterns that are recognized and have a response from the brain since born or during any stage of our life. The modeling of these mental instinctive patterns is important because they can help to understand the possible descriptions of the learning process proposed below, and other characteristics such as animal and human desire, curiosity, the freedom of thought, free will, etc. Studies about the cognitive role of emotions, motivation and intuition are among current works[3,4] supporting the relevance of instinctive mental patterns.

Summarizing, the instinctive mental patterns may be important both as patterns that define the initial conditions of the mind and as the influence on behavior throughout life. Bio-physically, mental instinctive patterns may be associated with innate neural-physiological states in the brain structure or to some patterns of brain preferred responses that have the ability to be activated even without learning or prior memorization.

**3 - Mental Patterns Memorization**
The brain has the mental capacity to record some patterns so that they are activated (recalled) more easily in the future processing. These patterns are stored as memory, and can be reactivated as a response or remembering. The memorization of external patterns in the brain is widely scientifically known and accepted since the experiments of Pavlov and Hebb to the modern experiments with memory.

The memory bio-physical background can be modeled as an increase in the likelihood, frequency or the intensity of recognition or activation of a mental pattern related to an external physical pattern detection, or self functioning pattern detection.

**4 - Mental Patterns Activation** (dynamics or processing)

Recognition and memorization of patterns by the brain can be seen among the most important properties of what we can call the processing or dynamics of patterns, i.e., the brain has mechanisms to work with patterns in several ways. Other mechanisms of the brain activation patterns that I propose are:

- **Repetition of mental patterns** (automatic activation)
- **Association of mental patterns** (an active mental pattern actives others mental patterns)

The repetition of mental patterns is a mechanism I propose that might be considered analogous to the concept of cpu clock frequency in digital computers, i.e., in some way the brain is constantly activating various mental instinctive patterns, memorized mental patterns or patterns from the contact with the environment. The repetition of patterns of mental activation as will describe, may be a basis for the ongoing process of thought, the repetition of patterns of desires, curiosities, comparisons, ratings, etc, and also may be a basis for a scientific definition of free will, self-determination, etc.

The bio-physical basis of the repetition of the mental patterns can be any mental pattern firing mechanism commonly studied at neurobiology researches and some theoretical and computational models. Perhaps this is the most questionable of the proposed concepts, but, any other neural activation mechanisms one can suggest, it will activate and process mental patterns in a similar manner. I adopted this concept due to its feasibility and to its modeling simplicity.

The association of mental patterns can be compared as the arc-reflex, or the activation of a particular sensorial, instinctive or memory pattern from or after the activation of some other mental pattern. The computer analogy is the IF X THEN Y instruction, if pattern X is active then it actives pattern Y, which can also be regarded as the biological basis of the formal logic.

The bio-physical basis of the association of mental patterns is the same as the pattern recognition. In fact the association of mental patterns can be viewed as the recognition of the association pattern of mental patterns. I use the association of mental patterns as an auxiliary modeling concept.

The terms, instinct, memory, activation, repetition and association of mental patterns are useful to a comprehensive mind modeling that I propose hereafter to describe the key mental functions.

**Easy Problems** - Basics Solutions of Theory: Solutions proposed for the easy problems of cognitive science:

**Baby Learning**
To start an example of the explanatory power of a theory of mind from the principles previously defined, I suggest a possible description of the initial learning of a baby. Clearly the description below can be enriched and corrected by specialists in children and animal learning, but to build an initial descriptive model it is a valuable start for a global comprehension of the theory capability.

Forewords - A basic tactile-muscular pattern is that from a certain age the babies hold almost everything that touches its hand. This pattern is one of the stimuli that babies have as one of the first tactile interactions. With this initial babies reaction we can propose a model of how babies will develop certain mental patterns of its motor and touch abilities. We can model it with several repetitions of innate tactile stimuli and instinctive motor reactions of the muscles of the hand and arm. The muscles of the hand and arm will create the memory of how to work together achieve optimization and memorize mental patterns of muscle self-control, speed, positioning, scope, stability, pain, discomfort and comfort, etc. The patterns of comfort and discomfort for example can be instinctive patterns associated with the pain or fatigue. Summarizing, the motor system learns and memorizes the best mental motor patterns that fit the instincts patterns.

So from the basic principles defined we can describe a learning mechanism as modeled below:

Learning Engine Model - the mental patterns involved are:
- Mental patterns of instinctive reflex of closing the hand after the tactile sensation of a solid object pressing the palm hand.
- Mental patterns of memorization of patterns of tactile stimulation of an object in the palm of your hand and also the movements thereafter.
- Repetition of mental patterns memorized in closing the palm of the hand, and then the arm movement, and subsequent exploration of possible variations of muscular control of tactile sensory and with further memorized mental patterns.
- In the process of repetition and exploration of mental patterns of muscular movement control, the mental patterns that create associations with instinctively mental patterns of stability, comfort, no pain, etc, are recognized and memorized as preferred by the instinctive patterns.

Thus, the mental patterns of optimized muscles movement of the hand are memorized for future now learned movements.

Even if this simplified baby learning model is generally in accordance with some recent studies[5,5], a more detailed technical discussion from the psycho-motor and neural-motor specialties is need for an accurate description. But, whatever can be the complexity of the patterns of the motor learning, a theory of mental patterns can build an explanatory modeling.

With models similar to this we can make a full description of the development of body equilibrium of the babies, the development of vision, hearing, the patterns of association between all this, and all the basic behavior of animals.

Words - language – in the same way, we can model the mental patterns in which sounds are linked to our basic instincts as crying and laughing, and then with external patterns as the baby food asking, attractive object asking, etc.

In short, what the community of research in cognitive science calls the easy problem, can be solved considering the brain and mind as a dynamic system or processing mental patterns similar to what I am proposing here.

**Hard Problems** - Solutions proposed for the hard problems of cognitive science:

Following are some possible definitions, and descriptive solutions for the so called hard problems with a model of mind over dynamical mental patterns. Here I will not suggest a possible detailed formulation, as the above Learning Engine Model, because, as we have the absence of experimental and theoretical consensus in the community, I will better propose definitions and descriptive solutions that this model can construct. But I assume and believe that readers can readily understand the application of the model and the proposed definitions and solutions, as I do, and as I easily develop them bellow:

a) **Learning**
Is the dynamics of memorization and optimization of sensory patterns, organic-metabolic responses, sensory-motor, etc, as described above.

b) **Thinking**
A proposal for definition and description of thinking from mental patterns is that the human brain is repeating the activation of mental patterns even without an external stimulus, it activates and associates memorized patterns. In more familiar language, the brain is constantly remembering, comparing, remembering the stages of learning, comparing the strategies of learning and problem-solving, which in this theory is a proposed definition of thought. In other words, repeating memorized mental patterns according to association patterns simulate the repetition of memorized possible problem solutions, and also increases the probability to solve other problems, thus outlining tactics well in advance with the mental processing, or simple, it is thinking.

For short, the concept of comparing used here can also be modeled as the recognition of the patterns with relates two or more patterns in a qualifying pattern.

Any other aspects involved in the thinking concept (also in the consciousness concept thereafter) as one streaming of reasoning, meaning that there is not apparently a manifestation of two simultaneous thoughts, or the aspect of mental dominance, meaning that the human been seems in some respects to be guided by mental patterns related to basic instincts, like animals, but the thoughts appears to exert a domination in our decision making and hence in our actions and reactions.

Also, the human language activity seems to be strict related to this thinking mechanism, such that the memorized mental patterns (concepts) have a language pattern association. Only a few verbal expressions of shock, amazement, surprise, joy, etc, seem to be associated with unconscious mental process.

c) **Selfness**
The patterns that define the body contours, which are always present to us in any situation (the sensory organs of an individual are always perceiving the physical limits of the individual itself), the patterns of reaction, feeling, emotions that are always present in a pattern related to the own body, and the analogous recognition of similar patterns at other bodies (other individuals), may be the definition of selfness (self perception), or self identification according to the theory I am proposing. This is in line with some recent studies[7,8].

d) **Attention, Self-determination, Free Will** - in this theory I propose that we have instinctive mental patterns that constantly give us patterns of desire, curiosity, necessity of critical thinking, etc, as mental patterns which always direct us to seek patterns everywhere, every when, i.e., in the pattern repeating process we are always thinking. The modeling of this aspect may be on the concepts of repetition of patterns and instinctive patterns.

Also, one possible explanation to the feeling of freedom of thought, in the sense that we are mentally unbounded and free to think at an appearing infinite different manners, could be that the number of possible mental patterns at the available to the human brain is infinitely more numerous than what one life span of mind activity could ever process. If it is true, it could be related to the seemingly endless increasing of the knowledge, culture and scientific development of the humanity, which is far beyond than we could expect for from a darwinian-genetic evolutionary perspective. Only information (optimized learned mental patterns) touched through the generations could accumulate the human cultural heritage.

e) **CONSCIOUSNESS**
The mental activities defined above (learning, thinking, self-perception and self-determination or free will, etc.) are mental patterns that are in constant mental activity in the life of human beings (property of repetition of mental patterns). In the act of thinking, our basic ability to recognize patterns of activity is also able to recognize the own mental activity of thinking. That is the thinking is also a pattern that in this model can be identified by the mechanism of recognition of mental patterns, memorized and used as a concept, tool or mental pattern of thought.

Then I propose a definition of a **conscious thinking as the patterns of mental activity that are recognized by the mind pattern recognition mechanism as a pattern of thought**. And also of one of the most intriguing aspect of the consciousness concept of self-consciousness[9], meaning that we are conscious to be conscious, can be defined as that we can recognize that we have the general pattern recognition process of conscious thinking.

It may seems like a circular argument, but it is not upon the assumption that the brain on identifying patterns of its own thinking is becoming conscious of its own condition of been a thinking mechanism (to be aware of itself consciousness).

**Representation of Mind**
As already mentioned, the correlation between the mental patterns with the external world function and self functioning patterns, and also the memory and dynamical properties of the mental patterns constitutes the representational space for this theory of mind.

**Qualia**
In this theory, the sense of quality of a certain concept, its evaluation and comparison with other concepts is modeled as the interaction and influence of instinctive patterns with environmentally learned or culturally developed patterns. So the qualia can be defined as the recognition of this influence of instinctive patterns over the thinking process.

## Other topics

**Difference between animal and human mind**
First, we see many similarities in cognitive model with respect to aspects of basic learning, perception and reaction for the human and animal mind.

What can be different in human and animal mind are that the human seems to have a more memory capacity, and as a consequence, more mental patterns available. More mental patterns available possibly allows more planning thinking activity with no loss of basic instinctive and life maintenance mental patterns, as it can not be the case for a limited brain capacity. This seems to be the case for some brain developed animals like apes and orcas that seems to have cultural behavior.

Another possible cause is that, for animals the instinctive mental patterns for basic life maintenance, as food search, may be more dominant for the mental processing than for us humans.

**Neurobiology**
We can suggest the interpretation that the behavioral and the pattern interaction involved in these experiments are related mental patterns. But this area of study is very specific and very careful experiments needs to be designed for doing so.

**Proposals for Artificial Minds**
There are many artificial systems, computational algorithms and hardware with sensory capacity to perform patterns recognition. Also are the standard models based on neural networks or traditional artificial intelligence has several applications. So the principles and the model discussed here can be applied as an interpretation and theoretical basis, and also be used to suggestions and predictions.

A prediction that can be readily proposed is that when the artificial systems pattern recognition becomes developed enough to perform integrated processing of pattern recognition, memory and association, it will be close to reproduce artificially other functions as higher human thinking.

A suggestion to the field of AI is the development of software to implement artificial minds to recognize and process patterns of bits, or patterns of pixels, etc. This is accessible with the current technology and could have the ability to memorize, associate, have artificial instincts, to learn, to memorize the learning, etc, which could be a first step in practical and theoretical research of the mind, and would be virtually the experimental confirmation of this theory.

**Natural, cognitive and Social Sciences**
This is also a candidate model to unify the human and social sciences in the scope of the natural sciences. Having a model of mind based on a natural physical mechanism, any science, culture or concept is ultimately explainable by the natural sciences.

**Philosophy**
The pattern recognition mechanism and the universe viewed as entities recognized by its patterns is a self consistent set of principle that can be a basis for a theory of everything. This is, can be a complete explanatory philosophy of the universe, once in this model every entity or concept that the human mind metalizes is patterns. Every definition of a concept implies that the set of patterns that define the concept must be defined. So the principle of pattern recognition is a candidate to be an ultimate philosophy. In short, I am proposing the following assertion: **if a pattern is recognized by a mental mechanism or system, then this pattern exists for the mind of that mechanism**.

**Adopted Theoretical Approach**
This is a descriptive theory of the mental process from basic natural observed principles. It is inspired by the same aspect that Darwin Theory of Evolution is a descriptive theory of the process of development and differentiation of species. Even if this theory of mind will not show correctly after tests, technical analysis and criticism, it still may have a similar importance to the theoretical study of mind as Lamarck theory was one of the first theories on natural grounds that were proposed to explain the life diversity.

Also, I propose that the pattern recognition principle is can be very adaptive, in the sense that any definitions and the descriptive modeling mechanisms different from the developed here can described by a pattern recognition model. I argue that it is due to the fact that the patterns of any definition or any model can be developed by a suitable pattern recognition mechanism.

**Conclusion**
The concepts of pattern recognition, memorization, instinct and processing permit us to build a comprehensive and conceptually structured descriptive theory of mind from physical basis. I propose to bring the actual important concept of pattern recognition to a status of a principle to understand the mind. Also, what makes this modeling approach promissory is the assumption that the patterns of any other modeling details and paths can be, in principle, enclosed to a pattern recognition principle.

The assumption that a pattern recognition system can recognize the patterns of its own cognitive activity, allow us to build a comprehensive definition of the concept of consciousness.

Finally, the area of the so called cognitive science is actually considered to not have an overall accepted scientific model[10]. This article brings a serious attempt to put the fundamental questions about the mind under scientific answers without the need of heterodox hypothesis.

**References and Notes**

1. *AP&P Attention Perception and Psychophysics* journal from the Psychonomic Society and *Psychophysiology* journal from the Society of Physiological Research among others.
2. *Pattern Recognition* journal from the Pattern Recognition Society and *IEEE Transactions on Pattern Analysis* and *Machine Intelligence* from the IEEE Computer Society among others.
3. Luiz Pessoa, How do emotion and motivation direct executive control?, *Trends in Cognitive Sciences*, Vol.13 No.4 (2009).
4. Fernand Gobet and Philippe Chassy, Expertise and Intuition: A Tale of Three Theories, *Minds & Machines*, (2009) 19:151–180.
5. Claes von Hofsten, An action perspective on motor development, *Trends in Cognitive Sciences* Volume 8, Issue 6, 1 June 2004, Pages 266-272
6. Jeffrey M. Zacks *, Shawn Kumar, Richard A. Abrams and Ritesh Mehta, Using movement and intentions to understand human activity, *Cognition* 112 (2009), Pages 201–216.
7. Bigna Lenggenhager, Tej Tadi, Thomas Metzinger and Olaf Blanke1, Video Ergo Sum: Manipulating BodilySelf-Consciousness, *Science* 317, 1096 (2007).
8. Alvin Goldman1 and Frederique de Vignemont2, Is social cognition embodied?, *Trends in Cognitive Sciences* Vol.13 No.4 2009
9. Van Gulick, Robert, Consciousness, *The Stanford Encyclopedia of Philosophy (Winter 2008 Edition)*, Edward N. Zalta (ed.), URL=<http://plato.stanford.edu/archives/win2008/entries/consciousness/>.
10. Greg Miller, What Is the Biological Basis of Consciousness?, *Science* Vol. 309. no. 5731, p. 79 (2005).